\documentclass[conference]{IEEEtran}
\IEEEoverridecommandlockouts
\usepackage{cite}
\usepackage{amsmath,amssymb,amsfonts}
\usepackage{algorithmic}
\usepackage{graphicx}
\usepackage{textcomp}
\usepackage{xcolor}
\def\BibTeX{{\rm B\kern-.05em{\sc i\kern-.025em b}\kern-.08em
    T\kern-.1667em\lower.7ex\hbox{E}\kern-.125emX}}

\usepackage{comment}
\usepackage{bm}
\usepackage{mathtools}
\usepackage{url}

\begin{document}

\title{Simple Domain Generalization Methods are\\Strong Baselines for Open Domain Generalization
\thanks{This work was partially supported by JSPS KAKENHI (JP23H00491, JP23H03466), JST PRESTO (JPMJPR2133), and NEDO (JPNP18002, JPNP20006).}
}

\author{\IEEEauthorblockN{1\textsuperscript{st} Masashi Noguchi}
\IEEEauthorblockA{
\textit{Graduate School of Environment and Information Sciences} \\
\textit{Yokohama National University}\\
Yokohama, Japan \\
noguchi.kdk.mm@gmail.com}
\and
\IEEEauthorblockN{2\textsuperscript{nd} Shinichi Shirakawa}
\IEEEauthorblockA{
\textit{Faculty of Environment and Information Sciences} \\
\textit{Yokohama National University}\\
Yokohama, Japan \\
shirakawa-shinichi-bg@ynu.ac.jp}
}

\maketitle

\begin{abstract}
In real-world applications, a machine learning model is required to handle an open-set recognition (OSR), where unknown classes appear during the inference, in addition to a domain shift, where the data distribution differs between the training and inference phases. Domain generalization (DG) aims to handle the domain shift situation where the target domain of the inference phase is inaccessible during the model training. Open domain generalization (ODG) considers DG and OSR. Domain-augmented meta-learning (DAML) is a method targeting ODG; however, it has a complicated learning process. By contrast, although various DG methods have been proposed, they have not been evaluated in ODG situations. In this study, we comprehensively evaluate the existing DG methods in ODG and show that the two simple DG methods, CORrelation ALignment (CORAL) and maximum mean discrepancy (MMD), are competitive with DAML in several cases. In addition, we propose simple extensions of CORAL and MMD by introducing the techniques used in DAML, such as ensemble learning and Dirichlet mixup data augmentation. The experimental evaluation demonstrates that the extended CORAL and MMD can perform comparably to DAML with lower computational costs. This suggests that the simple DG methods and their simple extensions are strong baselines for ODG.
\end{abstract}


\section{Introduction}
Machine learning has been significantly successful in various fields, such as image and speech recognition and natural language processing~\cite{sze2017efficient}. However, the performance of machine learning models is significantly degraded when the data distribution differs between the training and inference (testing) phases~\cite{hendrycks2018benchmarking}. This situation is called domain shift and often appears in real-world machine learning applications~\cite{quinonero2008dataset}, including computer vision tasks. The domains (i.e., the data distribution or characteristics) that can be accessed in model training are called the source domains. In contrast, the domains in the testing phase are called the target domains. Domain generalization (DG)~\cite{DGsurvey1,DGsurvey2} is an approach that addresses the domain shift. The DG methods only use the source domain data during the training phase and do not use the target domain data.

Open-set recognition (OSR)~\cite{osrsurvey2020,osrsurvey2021} is another requirement in real-world applications of the classification models. Although the class space, i.e., the number and contents of classes, is identical between the training and testing phases in the usual machine learning and DG situations, OSR deals with situations where the source and target class spaces differ. The objective of OSR is to detect unknown classes while correctly recognizing the known classes in the testing phase. Most OSR methods do not consider the domain shift between the training and testing phases.

Open domain generalization (ODG), introduced by \cite{DAML}, is the situation that considers DG and OSR. Although ODG is also frequently encountered in real-world applications, such as autonomous driving and medical assistance, the methods for ODG have not been fully developed. Domain-augmented meta-learning (DAML)~\cite{DAML} is a method that addresses ODG, but it has a complicated learning process and consists of several training components.

The work of \cite{gulrajani2020search} showed that the carefully implemented and tuned simple empirical risk minimization (ERM) method without any specific technique for DG outperforms many state-of-the-art DG methods. Furthermore, in OSR, ERM with several simple modifications, such as data augmentation, label smoothing, and the use of logits instead of softmax probabilities, has exhibited performance comparable to that of the more complex state-of-the-art OSR methods~\cite{closed-set-all-you-need}. The lesson from these studies is the importance of comprehensively evaluating simple baseline methods rather than developing complicated ones. Simple methods are generally advantageous with respect to computational cost, implementation, and ease of combining other techniques.

Because existing DG methods have not been thoroughly evaluated in ODG situations, we first comprehensively evaluate the existing DG method in ODG situations. We show that the performances of the simple DG methods, CORrelation ALignment (CORAL)~\cite{CORAL} and maximum mean discrepancy (MMD)~\cite{DGverMMD}, are better than that of other DG methods and are comparable to that of the state-of-the-art DAML on datasets where the effectiveness of data augmentation is low. Because CORAL and MMD are simple and easy to combine with other enhanced training methods, we introduced the techniques used in DAML, including ensemble learning, Dirichlet mixup (Dir-mixup) data augmentation, and knowledge distillation, into CORAL and MMD to improve their performances. The experimental result demonstrates that CORAL and MMD with Dir-mixup data augmentation achieve performances comparable to that of DAML on datasets where the data augmentation is effective. The contributions of this paper are as follows.
\begin{itemize}
    \item We comprehensively evaluate the DG methods in ODG.
    \item We propose the extended methods of CORAL and MMD by injecting the techniques used in DAML.
    \item Empirical evaluation reveals that simple DG methods and their extensions can be strong baselines for ODG, which would be useful for practitioners and further algorithm developments.
\end{itemize}
We note that our goal is not to develop sophisticated methods for ODG. We aim to find simple DG methods that compete with the state-of-the-art method. Exploring the potential of simple baseline methods is valuable for future algorithm development.

\section{Definitions of DG and ODG}

\subsection{Domain Generalization (DG)}
Let us consider classification problems and denote the input and class spaces as $\mathcal{X}$ and $C$, respectively. We have $M$ multiple labeled source datasets (domains), $\mathcal{D}_1, \dots, \mathcal{D}_M$, where $\mathcal{D}_s=\left\{\left(\mathbf{x}_s^i, \mathbf{y}_s^i\right)\right\}_{i=1}^{N_s}$, $(\mathbf{x}_s^i, \mathbf{y}_s^i) \in \mathcal{X} \times C$, and $N_s$ indicates the number of data in the $s$-th dataset. The goal of DG is to obtain a model that correctly predicts the known classes of unseen input data $\{ \mathbf{x}_t \}_{i=1}^{N_t}$ in a target domain $\mathcal{D}_t$ only using multiple source domains. In DG, the data distributions of the source and target domains $\mathcal{D}_1, \dots, \mathcal{D}_M, \mathcal{D}_t$ are different, but the class spaces are identical, i.e., $\mathbf{y}_s^i, \mathbf{y}_t^i \in C$. Note that the target domain data are inaccessible during model training.

\subsection{Open Domain Generalization (ODG)}
In ODG~\cite{DAML}, the class spaces of the source domains can be different, that is, $C_p \neq C_q$ for the $p$- and $q$-th source domains. In addition, the target domain has novel classes that do not belong to any source domain. Let $C=\bigcup_{k=1}^{M} C_k$ be the union of all source class spaces; then we denote the target class space $C_t = C \cup C_u$, where $C_u$ is the novel class space that first appears in the test phase. This discrepancy in the class spaces between the source and target domains is the difference between DG and ODG and the difficulty of ODG compared to DG. The goal of ODG is to obtain a model that correctly classifies the unseen target samples $\{ \mathbf{x}_t \}_{i=1}^{N_t}$ into $\left|C\right|+1$ classes, that is, classifying samples belonging to known classes $C$ into the correct class and treating samples belonging to novel classes $C_u$ as one unknown class. We note that, as in the case of DG, the data distributions of the source and target domains $\mathcal{D}_1, \dots, \mathcal{D}_M, \mathcal{D}_t$ are different, and the target domain data is inaccessible during model training. 

A similar scenario where input data in target domains are accessible is called open domain adaptation (ODA)~\cite{Saito_2018_ECCV}, which is a combination of domain adaptation and OSR. ODA is a more realistic problem than the general domain adaptation, but it assumes that target data are available during the training phase while they are inaccessible in ODG.

\section{Key Algorithms}

\subsection{CORAL and MMD} \label{ref DG}
CORAL~\cite{CORAL} and MMD~\cite{DGverMMD} are the classic DG methods, which introduce the additional term in the loss function.

CORAL uses the loss $\ell_{CORAL}$, which is the sum of the difference in the covariance matrices $C_{i}$ and $C_{j}$ of the features for each source domain. The loss $\ell_{C O R A L}$ is given by
\begin{equation}
\label{eq:DGCORAL}
\ell_{C O R A L}= \frac{1}{M^2} \sum_{1 \leq i \leq M} \sum_{1 \leq j \leq M} \left\|C_{i}-C_{j}\right\|_F^2 \enspace ,
\end{equation}
where $M$ is the number of source domains, $C_i$ represents the covariance matrix of features for the $i$-th domain, and $\|\cdot\|_F$ is the Frobenius norm.

MMD~\cite{DGverMMD} measures the distance between the training data distributions of $P^{i}_X$ and $P^{j}_X$ by the maximum mean discrepancy~\cite{MMD}. The loss $\ell_{MMD}$ in our domain generalization scenario is defined by
\begin{equation}
\label{eq:DGMMD}
\ell_{MMD} = 
\frac{1}{M^2} \sum_{1 \leq i \leq M} \sum_{1 \leq j \leq M} \text{MMD}(P^{i}_X, P^{j}_X)^{2} \enspace ,
\end{equation}
where $\text{MMD}(P^{i}_X, P^{j}_X)$ measures the distance between the feature distributions of the $i$- and $j$-th domains.

In model training, the cross-entropy loss with $\ell_{CORAL}$ or $\ell_{MMD}$ is minimized, and the covariance matrix $C_i$ in CORAL and the MMD function are calculated using mini-batch samples.

\subsection{Domain-augmented meta-learning (DAML)}
DAML is an ensemble method that prepares the same number of models as the source domains for training. Each model is trained in two stages by meta-learning using Dir-mixup data augmentation and knowledge distillation.

Dir-mixup creates a new sample by combining data from different domains, and the Dir-mixup data are used for training models to be robust against domain shifts. Let $(\mathbf{x}_k, \mathbf{y}_k)$ be the input/output data of the $k$-th source domain, $F_k$ and $G_k$ be the feature extractor and classifier for the $k$-th model, respectively, and $\mathbf{z}_k=F_k\left(\mathbf{x}_k\right)$ be the feature vector. Then, the Dir-mixup augmented data $(\mathbf{z}^{\mathrm{D\mathchar`-mix}}, \mathbf{y}^{\mathrm{D\mathchar`-mix}})$ is given by
\begin{gather}
  \boldsymbol{\lambda} \sim \operatorname{Dirichlet}(\boldsymbol{\alpha}) \label{eq:Dir-mixup alpha} \\ 
  \left(\mathbf{z}^{\mathrm{D\mathchar`-mix}}, \mathbf{y}^{\mathrm{D\mathchar`-mix}}\right)=\left(\sum_{k=1}^M \lambda^{(k)} \mathbf{z}_k, \sum_{k=1}^M \lambda^{(k)} \mathbf{y}_k\right) \enspace ,
  \label{eq:D-mixup}
\end{gather}
where $\bm{\lambda}$ is random variables sampled from the Dirichlet distribution parametrized by $\bm{\alpha} \in \mathbb{R}^M_+$. The parameter of $\bm{\alpha}$ is set according to the target model to train and the training stage.

Knowledge distillation is used to improve the performance of a target model by generating pseudo-labels from the outputs of other models. Knowledge distillation in DAML uses the weighted sum of the outputs of $M-1$ other models as the pseudo-label to train the target model. Let us consider that the $i$-th model is the target training model; then the pseudo-label $\mathbf{y}_i^{\mathrm{distill}}$ for the input $\mathbf{x}_i$ sampled from the $i$-th source domain is given by
\begin{gather}
    \boldsymbol{\lambda} \sim \operatorname{Dirichlet}(\boldsymbol{\alpha}^{\prime}) \label{eq:distill alpha}\\
    \mathbf{y}_i^{\mathrm{distill}}=\sum_{j=1}^{i-1} \lambda^{(j)} \mathcal{M}_j\left(\mathbf{x}_i\right)+\sum_{j=i+1}^M \lambda^{(j-1)} \mathcal{M}_j\left(\mathbf{x}_i\right) \enspace ,
    \label{eq:distill-label}
\end{gather}
where $\mathcal{M}_j=G_j \circ F_j$ denotes the composite function of the feature extractor and classifier of the $j$-th model, and $\bm{\alpha}' \in \mathbb{R}^{M-1}_+$ represents the Dirichlet distribution parameters, which are set as a vector with all elements $1$. 

In DAML, two types of loss functions, meta-training and meta-objective losses, are defined using the Dir-mixup samples and the knowledge-distilled pseudo-labels in addition to the original data. In model training, the model parameters of each domain's model are updated using two types of loss functions based on meta-learning. The detailed training procedure and the setting of loss functions can be found in \cite{DAML}.

In the testing phase, the prediction $\widehat{\mathbf{y}}_t$ is given by averaging the outputs of all models for a test sample $\mathbf{x}_t$ as follows:
\begin{equation}
\label{eq:inference}
\widehat{\mathbf{y}}_t=\frac{1}{M} \sum_{i=1}^M G_i\left(F_i\left(\mathbf{x}_t\right)\right) \enspace .
\end{equation}

The literature on DAML~\cite{DAML} assumed that multiple source domains are accessible during training. Although a method for open-set single-domain generalization with a single domain in training has recently been proposed~\cite{crossmatch, onering}, this study focuses on ODG with multiple source domains for training.
In \cite{DAML}, several DG methods have been evaluated in ODG as baselines. However, classic and simple DG methods, such as CORAL and MMD, have not been comprehensively evaluated in ODG.

\section{Application and Extension of DG Methods to ODG}
This section describes how a classification model trained by the DG methods detects unknown classes in \ref{sec:detect_unknown_class}. Subsequently, we introduce the extended methods of the simple DG methods, CORAL and MMD, by incorporating ensemble learning, Dir-mixup data extension, and knowledge distillation used in DAML in \ref{sec:extension-CORAL-MMD}.

\subsection{How to Detect Unknown Classes}
\label{sec:detect_unknown_class}
We simply detect unknown classes based on the class probability given by the model's outputs. Let $P(c \mid \mathbf{x})$ be the class probability of $c$ for a given data $\mathbf{x}$, then the predicted class $\hat{y}$ is determined by
\begin{equation}
\hat{y}=\left\{\begin{array}{ll}
\text{unknown} & \text {if} \;\; \forall c \ P(c \mid \mathbf{x})<\delta \\
\operatorname{argmax}_c P(c \mid \mathbf{x}) & \text {otherwise}
\end{array}\right. \enspace ,
\end{equation}
where $\delta$ is the threshold parameter, and the case of ``$\hat{y}=\text{unknown}$'' indicates that the input data $\mathbf{x}$ is judged to be an unknown class. A similar method is also used in DAML~\cite{DAML} to detect unknown classes.

\subsection{Extension of CORAL and MMD}
\label{sec:extension-CORAL-MMD}
CORAL and MMD are simple DG methods because they only add the specific loss term to minimize the distance between the feature distributions of source domains. We extend CORAL and MMD to improve the performance in ODG situations by injecting the components used in DAML. In DAML, two different loss functions using the Dir-mixup data augmentation and knowledge-distillated pseudo-labels are defined and minimized by alternative updates, which are complicated. We aim to adopt a simple way to introduce these techniques into CORAL and MMD.

\subsubsection{Extension by Ensemble Learning}
We first introduce ensemble learning across source domains into CORAL and MMD, called Ensemble-CORAL (E-CORAL) and Ensemble-MMD (E-MMD), respectively. We prepare $M$ neural network models $\{ \mathcal{M}_1, \dots, \mathcal{M}_M \}$ where the $i$-th model $\mathcal{M}_i = G_i \circ F_i$ corresponds to the $i$-th source domain, where $M$ is the number of source domains. Let us denote the source domains as $\mathcal{D} = \{\mathcal{D}_1, \dots \mathcal{D}_M \}$. The loss function for the $i$-th model is defined by
\begin{equation}
    \mathcal{L}_\mathrm{ens} (\mathcal{M}_i) = \sum_{j=1}^{M} w_j \mathcal{L}_{\mathrm{ce}} \left(\mathcal{D}_j, \mathcal{M}_i \right) + \mathcal{L}_\mathrm{reg} (\mathcal{D}, \mathcal{M}_i) \enspace ,
    \label{eq:Ensemble}
\end{equation}
where $\mathcal{D}_j$ indicates the $j$-th domain dataset, $\mathcal{L}_{\mathrm{ce}} (\mathcal{D}_j, \mathcal{M}_i)$ is defined as $\mathcal{L}_{\mathrm{ce}} (\mathcal{D}_j, \mathcal{M}_i) = \mathbb{E}_{(\mathbf{x}, \mathbf{y}) \sim \mathcal{D}_{j}} [\ell_\mathrm{ce} (\mathbf{y}, \mathcal{M}_{i}(\mathbf{x}))]$ with the cross entropy loss $\ell_\mathrm{ce}$. The term of $\mathcal{L}_\mathrm{reg}$ corresponds to the regularization terms of CORAL or MMD, that is, $\mathcal{L}_\mathrm{reg} = \ell_{CORAL}$ in Eq.~\eqref{eq:DGCORAL} in the case of E-CORAL and $\mathcal{L}_\mathrm{reg} = \ell_{MMD}$ in Eq.~\eqref{eq:DGMMD} in the case of E-MMD. The features extracted by $F_i$ in the model $\mathcal{M}_i$ are used to calculate the loss of CORAL and MMD. In addition, $w_j$ is a coefficient to balance the effect of the source domains. We specifically set $w_j = 3$ if $j = i$ and $w_j=1$ otherwise. This setting aims to learn the $i$-th model for the $i$-th domain while exploiting other domain datasets. We note that the training of E-CORAL and E-MMD can be parallelized across the models because the loss function is independent of the models.

\subsubsection{Extension by Dir-mixup Data Augmentation}
We further extend E-CORAL and E-MMD by incorporating Dir-mixup data augmentation, termed Ensemble-Dir-mixup-CORAL (EDir-CORAL) and Ensemble-Dir-mixup-MMD (EDir-MMD). In these extended methods, the loss using data samples generated by Dir-mixup data augmentation in Eq.~\eqref{eq:D-mixup} is added to Eq.~\eqref{eq:Ensemble}. The loss function for the model $\mathcal{M}_i$ is defined as
\begin{align}
    \mathcal{L}_\mathrm{Dir} (\mathcal{M}_i) = & \mathcal{L}_\mathrm{ens} (\mathcal{M}_i) \notag \\
    & + \mathbb{E}_{(\mathbf{z}, \mathbf{y}) \sim \mathcal{D}_{i}^{\mathrm{D\mathchar`-mix}}} [\ell_\mathrm{ce} (\mathbf{y}, G_{i}(\mathbf{z}))] \enspace ,
\end{align}
where $\mathcal{D}_{i}^{\mathrm{D\mathchar`-mix}}$ denotes the data distribution given by the Dir-mixup in Eq.~\eqref{eq:D-mixup} by setting the parameters of the Dirichlet distribution as $\alpha_k = 0.6$ (if $k = i$) and $\alpha_k = 0.2$ (if $k \neq i$).

\subsubsection{Extension by Knowledge Distillation}
We propose another option to extend E-CORAL and E-MMD using knowledge distillation. We term each method as Ensemble-Distill-CORAL (EDst-CORAL) and Ensemble-Distill-MMD (EDst-MMD). We add the loss using the knowledge-distilled labels given by Eq.~\eqref{eq:distill-label} to Eq.~\eqref{eq:Ensemble}. The loss function for the model $\mathcal{M}_i$ is defined as
\begin{align}
\mathcal{L}_\mathrm{dst} (\mathcal{M}_i) = & \mathcal{L}_\mathrm{ens} (\mathcal{M}_i) \notag \\
    & + \mathbb{E}_{(\mathbf{x}, \mathbf{y}) \sim \mathcal{D}_{i}^{\mathrm{distill}}} [\ell_\mathrm{ce} (\mathbf{y}, \mathcal{M}_{i}(\mathbf{x}))] \enspace ,
\end{align}
where $\mathcal{D}_{i}^{\mathrm{distill}}$ denotes the data distribution given by the knowledge distillation in Eq.~\eqref{eq:distill-label} using data samples from the $i$-th domain.

In the testing phase, the extended methods of CORAL and MMD adopt the same prediction as in Eq.~\eqref{eq:inference}.

\begin{figure*}[tb]
 \centering
 \includegraphics[width=0.85\linewidth]{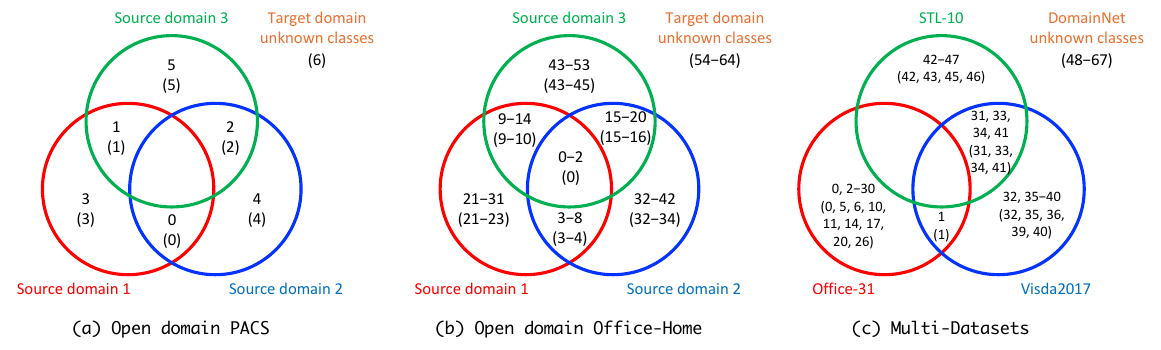}
 \caption{Illustration of the class space for each dataset. Indices without brackets show the distribution of classes among the source domains, while indices in brackets indicate the classes of the target domain.}
\label{fig:ven}
\end{figure*}

\section{Experiments and Results}
\subsection{Datasets}
We used three public datasets: PACS~\cite{PACS}, Office-Home~\cite{Office-Home}, and Multi-Dataets~\cite{DAML}. The PACS dataset is a seven-class image classification dataset; each class contains images from four domains: Photo (P), Art (A), Cartoon (C), and Sketch (S). Office-Home is a 65-class image classification dataset; each class contains images from four domains: Art (Ar), Clipart (Cl), Product (Pr), and Real-World (Rw). In the PACS and Office-Home datasets, the class space is the same for each domain; however, in the experiment, we used the open-domain PACS and open-domain Office-Home~\cite{DAML} datasets, which have different class spaces for each domain, as shown in Figures~\ref{fig:ven} (a) and (b). These settings are the same as in \cite{DAML}. We considered four experimental settings by picking up each domain as the target domain, and the remaining three domains are used for the source domains in alphabetical order. The correspondence between the source and target domains is shown in Table~\ref{tab:correspondence-PACS-Office-Home}.

We also used Multi-Datasets~\cite{DAML} consisting of several different datasets: Office-31~\cite{office31}, STL-10~\cite{STL-10}, VisDA2017~\cite{Visda2017}, and DomainNet~\cite{DomainNet}. As in the literature~\cite{DAML}, we use the Amazon domain from Office-31, STL-10, and the synthetic image domain from VisDA2017 as the source domains. The four DomainNet domains (Clipart, Painting, Real, and Sketch) were used as target domains.
The class space for each domain is shown in Figure~\ref{fig:ven} (c). We note that open-domain PACS and Multi-Datasets do not have a major class that appears in all three source domains.

\begin{table}[tb]

\centering
\caption{Correspondence between the source and target domains in the PACS and Office-Home datasets. We denote the source and target domain names as (source domains)-(target domain).}
\label{tab:correspondence-PACS-Office-Home}
\begin{tabular}{ccc}
\hline
  & PACS  & Office-Home \\ \hline
1 & (A, C, P)-(S) & (Ar, Cl, Pr)-(Rw)   \\
2 & (A, C, S)-(P) & (Ar, Cl, Rw)-(Pr)   \\
3 & (A, P, S)-(C) & (Ar, Pr, Rw)-(Cl)   \\
4 & (C, P, S)-(A) & (Cl, Pr, Rw)-(Ar)   \\ \hline
\end{tabular}
\end{table}

\begin{figure*}[tb]
\centering
\includegraphics[width=0.8\linewidth]{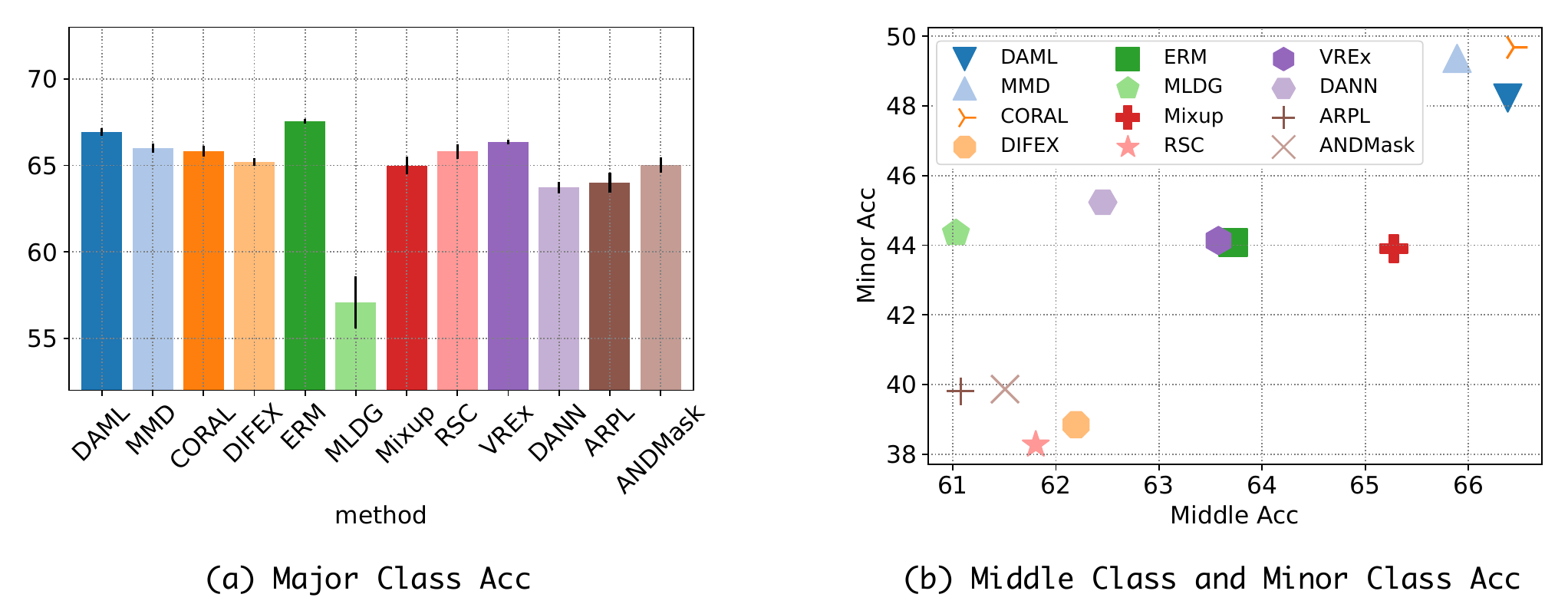}
\caption{(a) Average accuracy of the target data in major classes in open-domain Office-Home. The major class indicates that the class appeared in three source domains. (b) Average accuracy of the target data in middle and minor classes in the open-domain Office-Home. The middle and minor classes indicate the classes that appeared in two and one source domains, respectively.}
\label{fig:major-middle-minor-acc}
\end{figure*}

\subsection{Methods}
We evaluated a baseline method (ERM), an OSR method (ARPL), nine DG methods including CORAL and MMD, an ODG method (DAML), in addition to the extended methods of CORAL and MMD proposed in Section~\ref{sec:extension-CORAL-MMD}, as listed below.\footnote{Our code for the experiments is available at \url{https://github.com/shiralab/OpenDG-Eval}.}
\begin{itemize}
    \item {\bf Baseline method} (ERM~\cite{ERM}): A method without any handling for domain shift or OSR, which only minimizes the cross-entropy loss for source domain datasets.
    \item {\bf OSR method} (ARPL~\cite{ARPL}): A state-of-the-art method for OSR; however, it does not deal with domain shift.
    \item {\bf DG methods} (CORAL, MMD, ANDMASK~\cite{ANDMASK}, DANN~\cite{DANN}, MLDG~\cite{MLDG}, Mixup~\cite{zhang2018mixup}, RSC~\cite{RSC}, V-REx~\cite{V-REx}, and DIFEX~\cite{DIFEX}): These methods were used for the DG scenario.
    \item {\bf ODG method} (DAML~\cite{DAML}): A method proposed for ODG.
    \item {\bf Extension of CORAL and MMD} (E-CORAL, E-MMD, EDir-CORAL, EDir-MMD, EDst-CORAL, and EDst-MM): The methods introduced in Section~\ref{sec:extension-CORAL-MMD}.
\end{itemize}
We selected the DG methods from those evaluated in \cite{gulrajani2020search,DGsurvey1} and used the same implementation as in \cite{gulrajani2020search,DGsurvey1}. Among the DG methods used in our experiment, MLDG and RSC were evaluated in \cite{DAML} for ODG, while the other seven DG methods have not been evaluated yet for ODG.

The feature extractors used for each method were the pre-trained ResNet-18~\cite{ResNet} on ImageNet~\cite{ImageNet}. We used the early stopping strategy based on the validation accuracy and evaluated the model with the best accuracy during the training. The data augmentation is the same as that in the literature~\cite{gulrajani2020search}.
We list the common setting in Table~\ref{tab:common-settings}. Other hyperparameters of each algorithm are determined based on literature~\cite{gulrajani2020search,DGsurvey1}.

\begin{table}[tb]
\centering
\caption{Common experimental settings.}
\label{tab:common-settings}
\begin{tabular}{cc}
\hline
\textbf{Parameter} & \textbf{Value} \\
\hline
Trials      & 3         \\
Feature extractor     & ResNet-18~\cite{ResNet} \\
Optimizer & Momentum SGD      \\
Coefficient of inertia term & 0.9 \\
Batch size    & 32        \\
Num. of Training Epochs     & 100       \\
Early stopping epochs  & 10        \\
Learning rate       & 0.001     \\
\hline
\end{tabular}
\end{table}

We used two evaluation metrics: Accuracy (Acc) and H-score~\cite{h-score}, where Acc evaluates the accuracy of a known class classification and H-score equally evaluates known class classification and unknown class detection. The calculation method of H-score is the same as that in \cite{DAML}.

\begin{table*}[!t]
\centering 
\caption{Results of PACS dataset for each target domain under the open domain setting. Bold values represent the best value, underlined values represent the second-best value, and values in parentheses represent the standard deviation over three trials.}
\label{tab:PACS-DG}
\begin{tabular}{ccccccccccc}
\hline
         & \multicolumn{2}{c}{Art} & \multicolumn{2}{c}{Cartoon} & \multicolumn{2}{c}{Photo} & \multicolumn{2}{c}{Sketch} & \multicolumn{2}{c}{Avg.}          \\
Method   & Acc        & H-score      & Acc          & H-score        & Acc         & H-score       & Acc         & H-score        & Acc           & H-score            \\ \hline
ERM        &51.57           &43.82          &48.53          &41.82           &57.13          &60.00          &41.54          &36.61          &49.69 (1.17)           &45.56 (1.27) \\
\hline
ARPL       &53.47           &45.19          &52.86          &42.33           &56.70          &56.51          &46.55          &37.83          &52.39 (1.81)           &45.47 (1.27) \\
\hline
ANDMask    &50.51           &42.64          &\underline{57.31}  &\underline{48.45}   &57.35          &18.90          &47.29          & 40.29         &53.11 (0.85)           &37.57 (2.23) \\
DANN       &51.67           &42.64          &57.09          &47.24           &51.56          &43.58          &45.64          &38.16          &51.49 (2.68)           &42.90 (1.89) \\
DIFEX      &46.24           &40.38          &50.66          &44.13           &49.11          &10.84          &43.10          &40.41          &47.27 (0.90)           &33.94 (2.26) \\
MLDG       &48.53           &41.11          &44.45          &42.30           &61.04          &64.77          &40.97          &36.40          &48.75 (1.88)           &46.14 (0.53) \\
Mixup      &52.67           &42.59          &53.18          &44.93           &58.13          &58.63          &43.86          &37.38          &51.96 (1.29)           &45.88 (0.56) \\
RSC        &50.40           &36.54          &52.07          &46.72           &53.04          &48.48          &44.40          &37.57          &49.98 (0.99)           &42.33 (0.66) \\
VREx       &51.86           &43.90          &50.04          &42.38           &57.05          &59.01          &41.47          &35.72          &50.10 (1.34)           &45.25 (1.46) \\
CORAL      &\underline{53.72}   &\underline{47.24}  &55.89          &46.21           &\underline{71.13}  &\textbf{70.53} &51.95          &\underline{47.20}   &\underline{58.17} (1.96)   &\underline{52.79} (1.36) \\
MMD        &53.38           &44.26          &55.30          &45.96           &68.04          &\underline{69.79}  &\underline{54.24}  &\underline{47.20}   &57.74 (4.23)           &51.80 (1.64) \\
\hline
DAML       &\textbf{59.32}  &\textbf{51.62} &\textbf{66.70} &\textbf{53.25}  &\textbf{81.01} &63.68          &\textbf{62.97} &\textbf{53.30}  &\textbf{67.50} (1.80)  &\textbf{55.46} (4.31) \\
\hline
\end{tabular}
\end{table*}

\begin{table*}[!t]
\centering 
\caption{Results of Office-Home dataset for each target domain under the open domain setting. Bold values represent the best value, underlined values represent the second-best value, and values in parentheses represent the standard deviation over three trials.}
\label{tab:Office-Home-DG}
\begin{tabular}{ccccccccccc}
\hline
         & \multicolumn{2}{c}{Art} & \multicolumn{2}{c}{Clipart} & \multicolumn{2}{c}{Product} & \multicolumn{2}{c}{Real World} & \multicolumn{2}{c}{Avg.}         \\
Method   & Acc        & H-score      & Acc          & H-score        & Acc          & H-score        & Acc           & H-score          & Acc            & H-score          \\ \hline
ERM         &44.01               &43.84          &48.13             & 44.72         &56.32          & 54.20          &64.44           &57.75              &53.22 (0.74)   &50.13 (0.12) \\
\hline
ARPL        &40.84               &42.13          &45.04             & 42.73         &53.48          &52.31           &59.59          &54.59              &49.73 (0.63)   &47.94 (0.28) \\
\hline
ANDMask     &40.36            &41.10          &42.98                & 41.62         &54.32          & 53.38          &61.82          &55.01             &49.87 (0.72)   &47.77 (0.21) \\
DANN        &47.04              &45.26          &42.86              & 41.44         &56.43         & 51.74           &65.76           &58.47             &53.02 (0.25)   &49.23 (0.32) \\
DIFEX       &40.67             &40.93          &44.20               &43.22          &52.86         &50.84            &60.75           &54.67             &49.62 (0.21)   &47.41 (0.21) \\
MLDG        &44.06              &44.12          &44.33              & 46.23         &54.92           &53.29          &64.02          &58.49             &51.83 (0.58)   &50.53 (1.52) \\
Mixup       &46.09             &46.11            &46.60             &44.08          &57.14           &54.97          &65.34           &58.99            &53.79 (0.25)   & 51.03 (0.50) \\
RSC         &40.07               &40.11          &44.79             & 42.38         &52.91           & 53.62         &59.33          &55.31              &49.27 (0.10)   &47.85 (0.41) \\
VREx        &44.36              &44.59          &46.90              & 43.68         &56.86           & 54.43         &64.60           &57.84             &53.18 (0.80)   &50.13 (0.15) \\
CORAL       &\textbf{50.33}     &\textbf{47.28} &\underline{49.97}      &\underline{47.67}  &\textbf{60.49}  &\underline{56.15}  &\underline{68.47}  &\underline{61.94}      &\textbf{57.31} (0.70)   &\underline{53.26} (0.34) \\
MMD         &\underline{49.91}      &\underline{47.13}  &49.51              &46.82          &59.62           & 55.57         &\textbf{68.60}  &61.63              &\underline{56.91} (0.13)   &52.79 (0.58) \\
\hline
DAML        &47.64              &46.27          &\textbf{51.56}     &\textbf{49.41} &\underline{59.98}   &\textbf{56.97} &67.40           &\textbf{62.22}     &56.65 (0.89)   &\textbf{53.72} (0.43) \\
\hline
\end{tabular}
\end{table*}

\begin{table*}[!t]
\centering
\caption{Results of Multi-Datasets for each target domain. Bold values represent the best value, underlined values represent the second-best value, and values in parentheses represent the standard deviation over three trials.}
\label{tab:Multi-Dataset-DG}
\begin{tabular}{ccccccccccc}
\hline
         & \multicolumn{2}{c}{Clipart} & \multicolumn{2}{c}{Sketch} 
         & \multicolumn{2}{c}{Painting} & \multicolumn{2}{c}{Real}  & \multicolumn{2}{c}{Avg.}         \\
Method   & Acc        & H-score      & Acc          & H-score        & Acc          & H-score        & Acc           & H-score          & Acc            & H-score          \\ \hline
ERM    &35.09           &38.47          &28.89          &33.14          &\underline{48.75}  &\textbf{51.65}  &\textbf{67.68} &\textbf{64.77} &45.10 (1.13)            &47.01 (0.83) \\
\hline
ARPL   & 32.02          &34.62          &27.24          &31.14          &46.30          &49.12          &64.21          &61.72          &42.44 (0.65)            &44.15 (0.47) \\
\hline
ANDMask   & 35.33       & 37.73         &29.54          &31.20          &39.72          & 42.04         &60.21          &56.13          &41.20 (1.15)            &41.77 (1.07) \\
DANN    &34.43          &36.79          &27.97          &29.68          &43.02           &44.95         &60.72          &56.64          &41.54 (1.48)            &42.01 (1.16) \\
DIFEX    &30.27         &34.30          &27.45          &30.68          &38.49          &42.70          &60.13          &59.64          &39.08 (2.19)            &41.83 (1.73) \\
MLDG    &28.43          &32.78          &33.45          &\underline{37.75}  &41.88          &44.41          &55.01          &57.69          &39.69 (2.10)            &43.16 (2.47) \\
Mixup    &36.92         &37.55          &26.85          & 30.18         &44.75          &46.61          &66.15          &62.71          &43.67 (2.01)            &44.26 (2.61) \\
RSC    &31.12           & 33.91         &21.79          & 25.48         &39.42          &42.52          &62.29          &59.54          &38.65 (1.22)            &40.36 (1.01) \\
VREx    &36.11          & 38.44         &28.83          & 32.07         &\textbf{48.83} &\underline{50.95}  &\underline{67.58}  &64.29          &45.34 (0.61)            &46.44 (0.77) \\
CORAL    &\underline{40.80} &\underline{42.26}  &34.02          &\textbf{37.76} &45.53          &48.91          &65.98          &64.45          &\textbf{46.58} (1.20)   &\textbf{48.35} (0.89) \\
MMD    &39.17           &41.23          &\underline{34.25}  &37.72          &46.37          &49.48          &66.22          &\underline{64.54}  &\underline{46.50} (0.77)    &\underline{48.24} (0.82) \\
\hline
DAML    &\textbf{42.69} &\textbf{44.38} &\textbf{34.43} &37.38          &45.53          &47.85          &62.77          &63.00          &46.35 (2.38)            &48.15 (2.02) \\
\hline
\end{tabular}
\end{table*}

\begin{table*}[!t]
\centering
\caption{Comparison of the mean Acc and H-score values of CORAL, MMD, DAML, and their variants across all test domains for each dataset. The performances have been averaged for the four test domains for each dataset per trial, and the three-trial averages are shown. Bold numbers represent the best values, underlined numbers represent the second-best value, and values in parentheses represent the standard deviation over three trials.}
\label{tab:result-extended-method}
\begin{tabular}{ccccccc}
\hline
             & \multicolumn{2}{c}{PACS}            & \multicolumn{2}{c}{Office-Home}              & \multicolumn{2}{c}{Multi-Datasets}          \\

Method   & Acc        & H-score      & Acc          & H-score        & Acc          & H-score     \\
\hline
CORAL                 &58.17 (1.96)            &52.79 (1.36)              &57.31 (0.70)           &53.26 (0.34)              &46.58 (1.20)            &48.35 (0.89)  \\
E-CORAL               &59.50 (1.24)            &43.23 (3.97)              &\underline{58.15} (0.06) &53.46 (0.19)            &\textbf{48.38} (0.54)   &\underline{49.13} (0.62)  \\
EDir-CORAL            &\textbf{68.27} (1.87)   &53.86 (1.81)              &57.44 (0.18)           &\textbf{53.96} (0.23)     &46.55 (0.82)            &\textbf{49.43} (0.70)  \\
EDst-CORAL            &66.25 (1.86)            &53.10 (3.17)              &57.81 (0.98)           &\underline{53.86} (0.63)  &46.60 (0.91)            &48.34 (1.54)  \\
\hline
MMD                   &57.74 (4.23)            &51.80 (1.64)              &56.91 (0.13)           &52.79 (0.58)              &46.50 (0.77)            &48.24 (0.82)  \\
E-MMD                 &58.60 (0.57)            &49.08 (0.85)              &\textbf{58.61} (0.80)  &53.55 (0.27)              &\underline{47.47} (1.60) &47.60 (1.50)  \\
EDir-MMD              &\underline{67.52} (3.08)    &55.53 (2.56)          &57.10 (0.18)           &53.61 (0.45)              &45.78 (2.29)            &47.67 (1.80)  \\
EDst-MMD              &65.68 (3.10)            &53.32 (3.03)              &56.46 (0.37)           &53.21 (0.48)              &46.15 (0.48)            &47.92 (0.76)  \\
\hline
DAML                  &67.50 (1.80)            &55.46 (4.31)              &56.65 (0.89)           &53.72 (0.43)              &46.35 (2.38)            &48.15 (2.02) \\
DAML w/o Dir          &66.49 (0.93)            &\textbf{55.99} (0.72)     &57.34 (0.73)           &53.62 (0.49)              &46.51 (1.06)            &47.94 (0.83)  \\
DAML w/o Dst          &67.28 (2.60)            &\underline{55.57} (1.38)  &56.45 (0.31)           &53.34 (0.34)              &45.49 (2.43)            &48.23 (2.51)  \\
DAML w/o Dir-Dst      &64.91 (1.62)            &53.73 (1.05)              &57.80 (0.35)           &53.59 (0.50)              &46.51 (0.94)            &48.42 (1.07)  \\
\hline
\end{tabular}
\end{table*}

\subsection{Experimental Results and Discussion}
\subsubsection{Evaluation of Conventional DG Methods}
Tables~\ref{tab:PACS-DG}, \ref{tab:Office-Home-DG} and \ref{tab:Multi-Dataset-DG} show the mean Acc and H-score over three trials for each target domain of PACS, Office-Home, and Multi-Datasets, respectively. From Table~\ref{tab:PACS-DG} shows that DAML performs better than other methods for most target domains on open-domain PACS datasets; CORAL and MMD show the second-best performance. Tables~\ref{tab:Office-Home-DG} and \ref{tab:Multi-Dataset-DG}, which list the results of the open-domain Office-Home and Multi-Datasets, show that CORAL and MMD achieved comparable performance with DAML in both Acc and H-score. These results imply that the simple DG methods, CORAL and MMD, have the potential for ODG even compared to the complicated DAML. One possible reason for the superior performance of CORAL and MMD is that the other DG methods are too specialized for the DG problems and thus may be unsuitable for handling the different class spaces between domains.

Figure~\ref{fig:major-middle-minor-acc} shows the accuracy of the major, middle, and minor classes in open-domain Office-Home. The major, middle, and minor classes indicate the classes that appeared in the three, two, and one source domains, respectively. In the major class, DAML, MMD, and CORAL are not significantly better than the other methods and are inferior to the baseline method ERM. In contrast, DAML, MMD, and CORAL outperform the other methods in the middle and minor classes, suggesting that they perform well in ODG because they can predict classes with fewer domains better than the other methods.

\subsubsection{Evaluation of Extended CORAL and MMD}
Table~\ref{tab:result-extended-method} compares the average Acc and H-score of CORAL, MMD, DAML, and their variations for the target domains. In addition to the extended versions of CORAL and MMD, we evaluate the variations of DAML: DAML without Dir-mixup for meta-training and meta-objective losses (DAML w/o Dir), DAML without knowledge distillation (DAML w/o Dst), and DAML without both Dir-mixup and knowledge distillation (DAML w/o Dir-Dst).

For open-domain PACS, EDir-CORAL and EDir-MMD have improved performances, which are comparable to that of DAML. This result indicates that the Dir-mixup data augmentation is effective for both CORAL and MMD in open-domain PACS. By contrast, DAML, DAML w/o Dir, and DAML w/o Dst exhibit similar performance, and DAML w/o Dir-Dst has lower performance than that of the other variants. This implies that one of the Dir-mixup and knowledge distillation is needed to ensure the DAML performance in the open-domain PACS. This observation also holds for our proposed CORAL and MMD variants.

In the open-domain Office-Home and Multi-Datasets, the Dir-mixup data augmentation shows slight improvement for CORAL and MMD. This might be because data augmentation is not effective for Office-Home and Multi-Datasets. The experimental result shows that simple extensions of CORAL and MMD can enhance the performance and achieve performances comparable to or better than that of DAML.

Table~\ref{tab:source-domain-accuracy} shows the test performance of each method in the source domains, where no domain shift occurs and only known classes appear. Evaluating the methods on source domains is important because we may not know whether the domain shift will occur in a practical use case. Extended versions of CORAL and MMD show good performances for all datasets. In particular, E-CORAL and E-MMD exhibit the best performance and outperform ERM. Therefore, simple ensemble learning will be effective for small or no domain shifts.

\begin{table}[tb]
\centering
\caption{Comparison of the mean Acc for each dataset on the source domains over three trials. For each trial, the average ACC for three source domains over four experimental settings is calculated. Bold numbers represent the best value, underlined numbers represent the second-best value, and values in parentheses represent the standard deviation over three trials.}
\label{tab:source-domain-accuracy}
\scalebox{1.0}{
\begin{tabular}{cccc}
\hline
Method             & PACS            & Office-Home              & Multi-Datasets          \\ \hline
ERM                   &97.33 (0.19)             &83.86 (0.30)              &93.30 (0.52)                \\
\hline
CORAL                 &90.35 (1.65)             &81.56 (0.51)              &89.86 (0.52)           \\
E-CORAL               &\underline{97.48} (0.12) &\underline{84.77} (0.26)  &\textbf{94.41} (0.07)   \\
EDir-CORAL            &92.97 (0.22)             &83.79 (0.41)              &89.93 (0.51)                \\
EDst-CORAL            &92.64 (0.57)             &83.64 (0.46)              &90.06 (0.52)                \\
\hline
MMD                   &89.89 (1.74)             &81.30 (0.55)              &89.50 (0.35)                   \\
E-MMD                 &\textbf{97.59} (0.22)    &\textbf{85.47} (0.26)     &\underline{94.30} (0.34)  \\
EDir-MMD              &92.77 (0.72)             &83.65 (0.41)              &89.93 (0.91)      \\
EDst-MMD              &92.65 (0.36)             &83.40 (0.46)              &89.77 (0.31)      \\
\hline
DAML                  &94.05 (0.63)             &83.84 (0.04)              &89.11 (0.23)      \\
DAML w/o Dir          &93.40 (0.05)             &83.71 (0.16)              &90.03 (0.87) \\
DAML w/o Dst          &94.23 (0.63)             &83.55 (0.12)              &89.78 (0.43) \\
DAML w/o Dir-Dst      &93.16 (0.28)             &83.42 (0.39)              &90.60 (0.62) \\
\hline
\end{tabular}
}
\end{table}

\begin{table}[tb]
\centering
\caption{Comparison of training times (in seconds) per epoch on one GPU. Bold numbers represent the best value, underlined numbers represent the second-best value, and the values in parentheses represent the standard deviation.}
\label{tab:computationel-cost}
\scalebox{1.0}{
\begin{tabular}{ccccccc}
\hline
Method             & PACS       & Office-Home     & Multi-Datasets    \\
\hline
ERM                   &\textbf{6.368} (0.007) &\textbf{15.92} (0.04)    &\textbf{6.424} (0.014)        \\
\hline
CORAL                 &10.29 (0.07)         &16.76 (0.05)               &\underline{10.14} (1.05)        \\
E-CORAL               &13.88 (0.02)         &17.59 (0.17)               &13.95 (0.02)        \\
EDir-CORAL            &17.33 (0.02)         &18.46 (0.26)               &17.40 (0.05)        \\
EDst-CORAL            &17.23 (0.01)         &18.61 (0.35)               &17.30 (0.10) \\
\hline
MMD                   &\underline{10.21} (0.12) &\underline{16.60} (0.14) &10.36 (0.39)        \\
E-MMD                 &14.39 (0.01)         &17.63 (0.19)               &14.45 (0.03)        \\
EDir-MMD              &19.22 (0.07)         &19.34 (0.11)               &19.24 (0.05)        \\
EDst-MMD              &19.05 (0.02)         &19.43 (0.21)               &19.01 (0.04) \\
\hline
DAML                  &32.86 (0.08)         &32.96 (0.05)               &33.08 (0.13)        \\
DAML w/o Dir          &24.85 (0.12)         &24.81 (0.11)               &24.79 (0.18) \\
DAML w/o Dst          &32.85 (0.06)         &32.92 (0.07)               &32.81 (0.09) \\
DAML w/o Dir-Dst      &24.56 (0.15)         &24.64 (0.10)               &24.54 (0.25) \\
\hline
\end{tabular}
}
\end{table}

\begin{figure}[tb]
 \centering
 \includegraphics[scale=0.45]{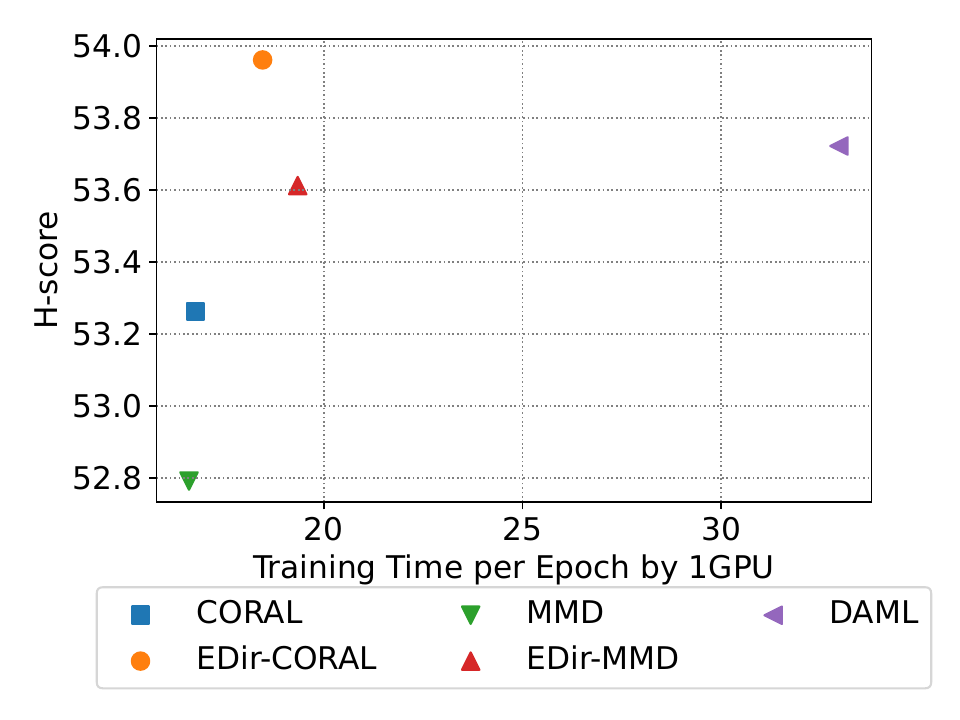}
 \caption{Scatterplot of H-score and training time per epoch (in seconds) for each method on the open-domain Office-Home.}
\label{fig:h-score-vs-time}
\end{figure}

\subsubsection{Computational Cost}
Table~\ref{tab:computationel-cost} compares the average training time per epoch of CORAL, MMD, and DAML variations. The model training uses an NVIDIA A100 (80GB) GPU. The training times of the variations of CORAL and MMD were less than that of DAML. This is because our CORAL and MMD variations update the model parameters by minimizing one loss function, whereas DAML alternatively minimizes two types of loss functions.
Note that, unlike DAML, E-CORAL and E-MMD can train each model in parallel, reducing the training time using more GPUs to the same level as that with CORAL and MMD. Figure~\ref{fig:h-score-vs-time} summarizes the relationship between the H-score for the target domain and training time on the open-domain Office-Home. We observe that EDir-CORAL and EDir-MMD achieve H-scores that are comparable to that of DAML at a lower training time cost than that of DAML.

Regarding the inference time, we note that those of the ensemble versions of CORAL and MMD are almost the same as that of DAML. However, simple CORAL and MMD are $M$ times faster than DAML in the inference stage, where $M$ is the number of source domains.

\subsubsection{Hyperparameter sensitivity}
\begin{figure}[tb]
\begin{center}
    \includegraphics[width=0.99\linewidth]{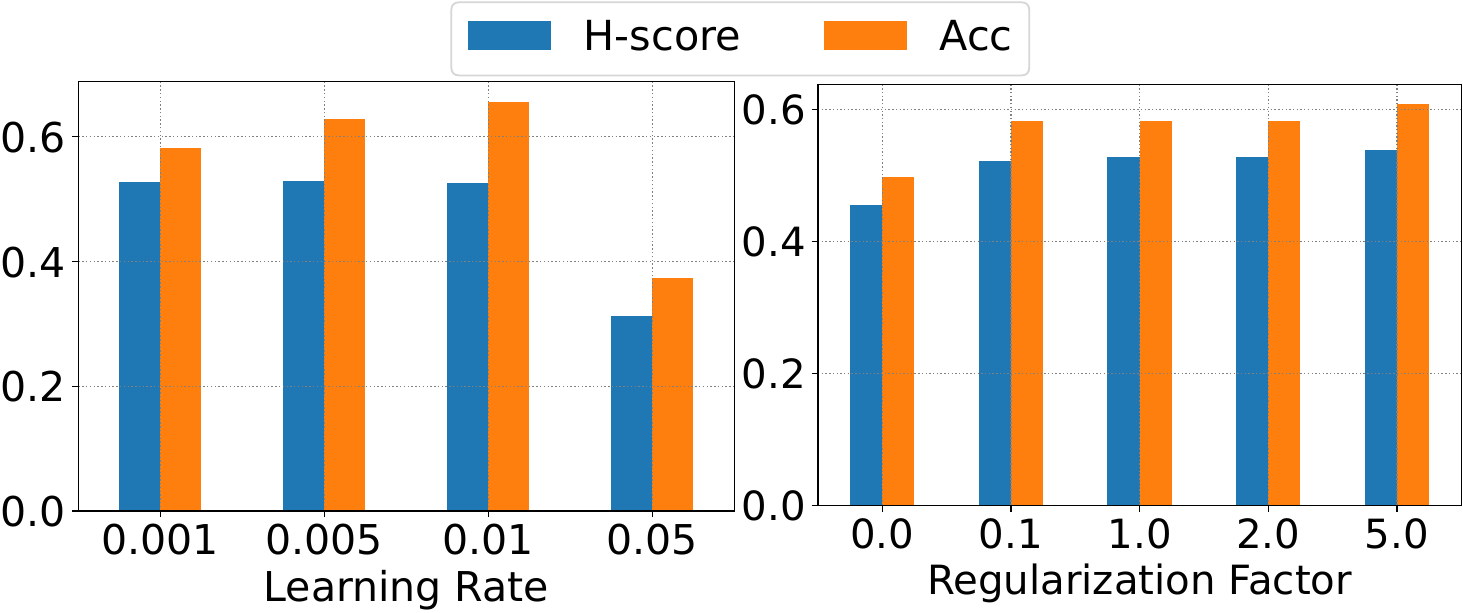}
\end{center}
   \caption{Hyperparameter sensitivity of CORAL on the PACS dataset (left: learning rate, right: coefficient of CORAL's loss).}
\label{fig:hp}
\end{figure}

The above experiments were conducted using fixed default hyperparameters. Although hyperparameter tuning is important to maximize the performance of machine learning models, it is not trivial to prepare the validation dataset and the metric for hyperparameter tuning because the validation performance for both known and unknown classes in the test domain should be considered in ODG. Therefore, we evaluated the existing methods with the default settings of public codes. However, as an example, we show the influence of the learning rate and the coefficient of CORAL's loss in Figure \ref{fig:hp}. Tuning the learning rate (default: 0.001) may improve the accuracy of CORAL, while the coefficient of CORAL's loss (default: 1.0) is not much sensitive to performance if it is greater than 0.1.

\section{Conclusion}
In this paper, we have comprehensively evaluated the DG methods in the ODG setting. We found that the classic DG methods, CORAL and MMD, are comparable to DAML proposed for ODG in two datasets out of three. We have also proposed and evaluated the extended versions of CORAL and MMD injected with the techniques used in DAML. We have shown that EDir-CORAL and EDir-MMD can reach performances comparable with that of DAML on all datasets with lower training costs. We expect that this work will be exploited as the baseline for the further development of the ODG methods. A possible future work is to evaluate the DG methods, including our extended versions, in the ODG situations of other tasks, such as speech and natural language processing. Theoretical analysis to understand why simple methods work well is another important future work.

\bibliographystyle{IEEEtran}
\bibliography{IEEEabrv,bibliography}

\clearpage
\appendix

\subsection{Detailed Experimental Setting}
We list hyperparameters of each algorithm in Table~\ref{tab:hyperparam}. We determined the hyperparameter values based on literature~\cite{gulrajani2020search,DGsurvey1}.

\begin{table}[tb]
\centering
\caption{Hyperparameters of each algorithm, which is set based on literature~\cite{gulrajani2020search,DGsurvey1}.}
\label{tab:hyperparam}
\begin{tabular}{ccc}
\hline
\textbf{Algorithm} & \textbf{Parameter}                                                                                                                                  & \textbf{Value}                                                                                 \\ \hline
ARPL       & \begin{tabular}[c]{@{}c@{}}lambda\\ temperature\\ margin\end{tabular}                                                                   & \begin{tabular}[c]{@{}c@{}}0.1\\ 1.0\\ 1.0\end{tabular}                                  \\ \hline
ANDMask    & tau                                                                                                                                         & 1.0                                                                                      \\ \hline
DANN       & alpha                                                                                                                                       & 1.0                                                                                      \\ \hline
DIFEX      & \begin{tabular}[c]{@{}c@{}}beta\\ dist type\end{tabular}                                                                                     & \begin{tabular}[c]{@{}c@{}}1.0\\ 2-norm\end{tabular}                                     \\ \hline
MLDG       & beta                                                                                                                                        & 1.0                                                                                      \\ \hline
Mixup      & alpha                                                                                                                                       & 0.2                                                                                    \\ \hline
RSC        & \begin{tabular}[c]{@{}c@{}}feature drop factor\\ batch drop factor\end{tabular}                                                                   & \begin{tabular}[c]{@{}c@{}}0.33\\ 0.33\end{tabular}                                    \\ \hline
VREx       & \begin{tabular}[c]{@{}c@{}}warmup iterations\\ lambda\end{tabular}                                                                                & \begin{tabular}[c]{@{}c@{}}500\\ 1.0\end{tabular}                                        \\ \hline
CORAL      & gamma                                                                                                                                       & 1.0                                                                                      \\ \hline
MMD        & gamma                                                                                                                                       & 1.0                                                                                      \\ \hline
DAML       & \begin{tabular}[c]{@{}c@{}}softmax temperature\\ trade\\ trade2\\ trade3\\ trade4\\ max alpha\\ min alpha\\ meta step size\end{tabular} & \begin{tabular}[c]{@{}c@{}}2.0\\ 3.0\\ 1.0\\ 1.0\\ 3.0\\ 0.6\\ 0.2\\ 0.01\end{tabular} \\ \hline
\end{tabular}
\end{table}

\subsection{Detailed Experimental Results}
In this section, we describe the accuracy of the middle and minor classes in open-domain PACS and Multi-Datasets. In addition, we show an evaluation of CORAL, MMD, DAML, and their variations in the target domain. Finally, we report the computational cost of CORAL, MMD, DAML, and their variations.

\subsubsection{Accuracy of Middle and Minor Classes in Open-Domain PACS and Multi-Datasets}
\begin{figure*}[tb]
 \centering
 \includegraphics[scale=0.35]{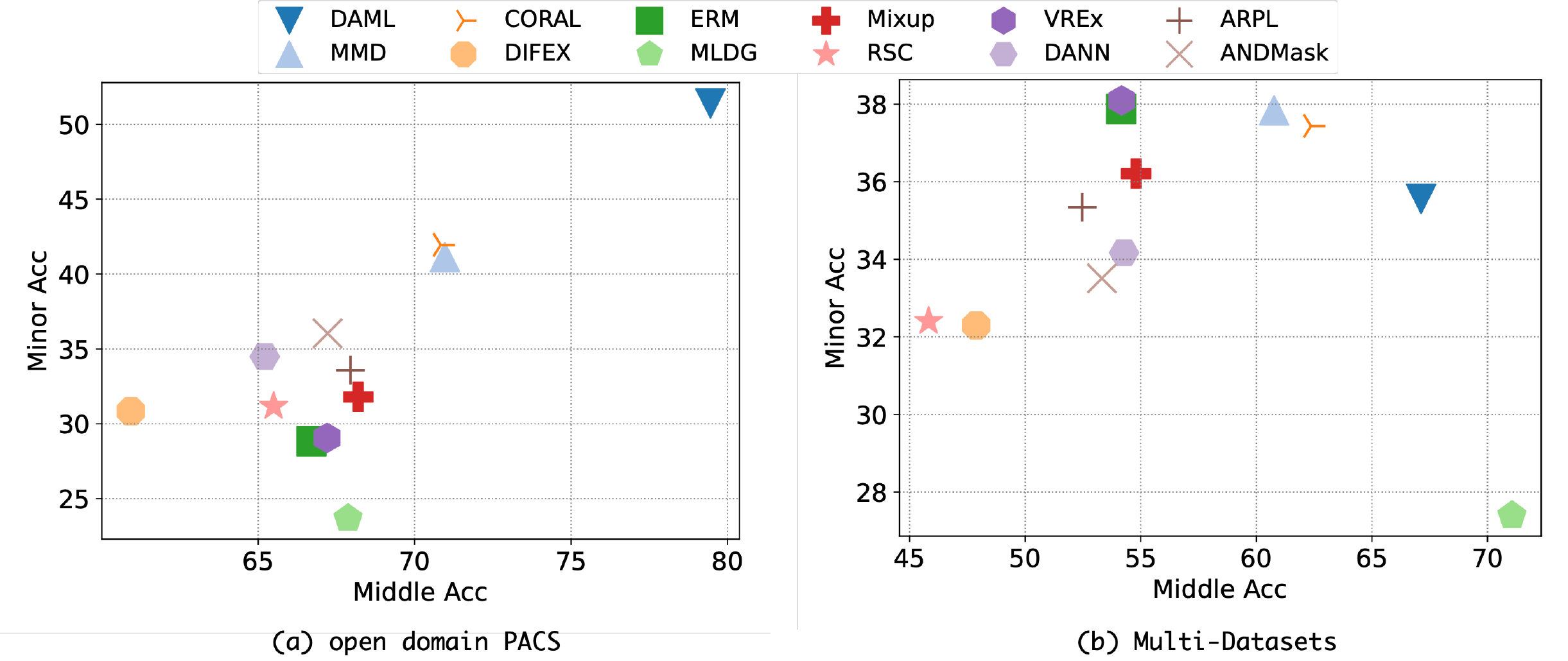}
 \caption{Scatterplot of middle and minor class Acc for (a) open-domain PACS and (b) Multi-Datasets.}
\label{fig:middle-minor-acc}
\end{figure*}

Figure~\ref{fig:middle-minor-acc} shows the accuracy of the middle and minor classes in open-domain PACS and Multi-Datasets with the existing methods. The middle and minor classes indicate the classes that appeared in two and one source domains, respectively.
In open-domain PACS, DAML, MMD, and CORAL outperform the other methods in the middle and minor classes. In Multi-Datasets, although DAML, MMD, and CORAL outperform the other methods except for MLDG in the middle class, they are not much superior to other methods in the minor class. Note that open-domain PACS and Multi-Datasets do not have a major class.

\subsubsection{Evaluation of CORAL, MMD, DAML, and Their Variations on Target Domain}
Tables~\ref{tab:PACS-result}, \ref{tab:Office-Home-result}, and \ref{tab:MultiDataSet-result} show the mean Acc and H-score over three trials by CORAL, MMD, DAML, and their variations for each target domain of PACS, Office-Home, and Multi-Datasets. From Table~\ref{tab:PACS-result}, we observe that in open-domain PACS, EDir-CORAL and EDir-MMD outperform CORAL and MMD for any target domain, showing the effectiveness of Dir-mixup data augmentation. On the other hand, Tables~\ref{tab:Office-Home-result} and \ref{tab:MultiDataSet-result} show that EDir-CORAL and EDir-MMD perform slightly better or similar for most target domains compared to CORAL and MMD.

\begin{table*}[tb]
\centering
\caption{Results of CORAL, MMD, DAML, and their variations for each target domain in the PACS dataset under the open-domain setting. Bold values represent the best value, underlined values represent the second-best value, and values in parentheses represent the standard deviation over three trials.}
\label{tab:PACS-result}
\scalebox{1.00}{
\begin{tabular}{ccccccccccc}
\hline
         & \multicolumn{2}{c}{Art} & \multicolumn{2}{c}{Cartoon} & \multicolumn{2}{c}{Photo} & \multicolumn{2}{c}{Sketch} & \multicolumn{2}{c}{Avg.}          \\
Method   & Acc        & H-score      & Acc          & H-score        & Acc         & H-score       & Acc         & H-score        & Acc           & H-score            \\ \hline
CORAL             &53.72           &47.24             &55.89          &46.21           &71.13               &\underline{70.53} &51.95          &47.20          &58.17 (1.96)           &52.79 (1.36) \\
E-CORAL           &59.16           &47.72             &59.62          &49.59           &63.41               &33.29          &55.83          &42.34          &59.50 (1.24)           &43.23 (3.97) \\
EDir-CORAL        &58.66           &49.67             &\underline{67.18} &\underline{53.31} &\underline{83.25} &60.30       &\underline{63.98}  &52.15          &\textbf{68.27} (1.87)  &53.86 (1.81) \\
EDst-CORAL        &\textbf{61.78}  &48.74             &64.81          &48.43           &80.02               &64.98          &58.38          &50.24          &66.25 (1.86)           &53.10 (3.17) \\
\hline
MMD               &53.38           &44.26             &55.30          &45.96           &68.04               &69.79          &54.24          &47.20          &57.74 (4.23)           &51.80 (1.64) \\
E-MMD             &57.55           &48.23             &57.33          &46.33           &64.43               &59.48          &55.11          &42.26          &58.60 (0.57)           &49.08 (0.85) \\
EDir-MMD          &59.56           &49.79             &66.40          &51.35           &79.59               &67.49          &\textbf{64.53} &\textbf{53.51} &\underline{67.52} (3.08)   &55.53 (2.56) \\
EDst-MMD          &59.45           &48.67             &62.97          &46.71           &78.73               &64.86          &61.59          &53.05          &65.68 (3.10)           &53.32 (3.03) \\
\hline
DAML              &59.32           &\underline{51.62} &66.70          &53.25           &81.01               &63.68          &62.97          &\underline{53.30}  &67.50 (1.80)           &55.46 (4.31) \\
DAML w/o Dir      &\underline{61.70} &51.44           &64.65          &52.44           &\textbf{83.38}      &\textbf{71.61} &56.22          &48.49          &66.49 (0.93)           &\textbf{55.99} (0.72) \\
DAML w/o Dst      &60.43           &51.08             &\textbf{67.20} &\textbf{53.47}  &80.42               &67.73          &61.08          &50.00          &67.28 (2.60)           &\underline{55.57} (1.38) \\
DAML w/o Dir-Dst  &61.01           &\textbf{52.07}    &63.78          &48.24           &83.14               &69.89          &51.71          &44.74          &64.91 (1.62)           &53.73 (1.05) \\
\hline
\end{tabular}
}
\end{table*}

\begin{table*}[tb]
\centering
\caption{Results of CORAL, MMD, DAML, and their variations for each target domain in the Office-Home dataset under the open-domain setting. Bold values represent the best value, underlined values represent the second-best value, and values in parentheses represent the standard deviation over three trials.}
\label{tab:Office-Home-result}
\scalebox{1.00}{
\begin{tabular}{ccccccccccc}
\hline
         & \multicolumn{2}{c}{Art} & \multicolumn{2}{c}{Clipart} & \multicolumn{2}{c}{Product} & \multicolumn{2}{c}{Real World} & \multicolumn{2}{c}{Avg.}         \\
Method           & Acc           & H-score       & Acc           & H-score        & Acc           & H-score         & Acc           & H-score       & Acc                  & H-score          \\ \hline
CORAL            &50.33          &\underline{47.28}  &49.97          &47.67           &60.49          &56.15            &68.47          &61.94          &57.31 (0.70)           &53.26 (0.34) \\
E-CORAL          &\underline{50.56}  &46.92          &51.77          &47.37           &60.78          &57.30            &\underline{69.50}  &62.27          &\underline{58.15} (0.06)   &53.46 (0.19) \\
EDir-CORAL       &48.24          &\textbf{47.40} &\underline{53.03}  &48.70           &59.62          &56.93            &68.89          &\underline{62.80}  &57.44 (0.18)           &\textbf{53.96} (0.23)\\
EDst-CORAL       &47.76          &46.33          &\textbf{54.23} &\textbf{51.29}  &59.76          &55.98            &68.41          &61.85          &57.81 (0.98)           &\underline{53.86} (0.63)\\
\hline
MMD              &49.91          &47.13          &49.51          &46.82           &59.62          &55.57            &68.60          &61.63          &56.91 (0.13)           &52.79 (0.58) \\
E-MMD            &\textbf{51.28} &47.09          &51.22          &47.31           &\textbf{61.42} &57.19            &\textbf{70.51} &62.64          &\textbf{58.61} (0.80)  &53.55 (0.27) \\
EDir-MMD         &46.33          &45.78          &52.63          &48.76           &\underline{61.34}  &\underline{57.55}     &68.12          &62.34         &57.10 (0.18)           &53.61 (0.45) \\
EDst-MMD         &47.10          &45.76          &51.87          &49.03           &59.96          &56.01            &66.92          &62.04          &56.46 (0.37)           &53.21 (0.48)\\
\hline
DAML             &47.64          &46.27          &51.56          &\underline{49.41}   &59.98          &56.97            &67.40          &62.22          &56.65 (0.89)           &53.72 (0.43) \\
DAML w/o Dir     &47.76          &46.18          &52.39          &49.08           &60.13          &56.37            &69.09          &\textbf{62.84} &57.34 (0.73)           &53.62 (0.49)\\
DAML w/o Dst     &45.97          &45.67          &52.35          &48.00           &60.07          &\textbf{58.37}   &67.41          &61.31          &56.45 (0.31)           &53.34 (0.34)\\
DAML w/o Dir-Dst &49.85          &47.24          &51.62          &48.43           &60.38          &56.30            &69.38          &62.40          &57.80 (0.35)           &53.59 (0.50)\\
\hline
\end{tabular}
}
\end{table*}

\begin{table*}[tb]
\centering
\caption{Results of CORAL, MMD, DAML, and their variations for each target domain in Multi-Datasets. Bold values represent the best value, underlined values represent the second-best value, and values in parentheses represent the standard deviation over three trials.}
\label{tab:MultiDataSet-result}
\scalebox{1.00}{
\begin{tabular}{ccccccccccc}
\hline
         & \multicolumn{2}{c}{Clipart} & \multicolumn{2}{c}{Sketch} 
         & \multicolumn{2}{c}{Painting} & \multicolumn{2}{c}{Real}  & \multicolumn{2}{c}{Avg.}         \\
Method   & Acc        & H-score      & Acc          & H-score        & Acc          & H-score        & Acc           & H-score          & Acc            & H-score          \\ \hline
CORAL               & 40.80         &42.26          &34.02          &37.76 &45.53          &48.91          &65.98          &64.45          &46.58(1.20)            &48.35 (0.89) \\
E-CORAL             &40.97          &42.41          &33.67          &35.94          &\textbf{49.65} &\textbf{52.15} &\underline{69.22}  &\textbf{66.04} &\textbf{48.38} (0.54)   &\underline{49.13} (0.62) \\
EDir-CORAL          &\underline{41.42}  &\textbf{44.46} &\textbf{37.72} &\textbf{41.03} &44.69          &48.99          &62.37          &63.24          &46.55 (0.82)            &\textbf{49.43} (0.70) \\
EDst-CORAL          &41.99          &44.59          &35.29          &37.25          &44.19          &47.38          &64.95          &64.15          &46.60 (0.91)            &48.34 (1.54) \\
\hline
MMD                 &39.17          &41.23          &34.25          &37.72          &46.37          &49.48          &66.22          &64.54          &46.50 (0.77)            &48.24 (0.82) \\
E-MMD               &39.70          &40.45          &31.95          &33.90          &\underline{48.53}  &\underline{50.17}  &\textbf{69.70} &\underline{65.89}  &\underline{47.47} (1.60)    &47.60 (1.50) \\
EDir-MMD            &40.27          &42.29          &35.56          &37.97          &44.00          &47.35          &63.31          &63.08          &45.78 (2.29)            &47.67 (1.80) \\
EDst-MMD            &40.80          &43.68          &35.67          &37.32          &43.43          &46.70          &64.71          &63.99          &46.15 (0.48)            &47.92 (0.76) \\
\hline
DAML                &\textbf{42.69} &\underline{44.38}  &34.43          &37.38          &45.53          &47.85          &62.77          &63.00          &46.35 (2.38)            &48.15 (2.02) \\
DAML w/o Dir        &40.89          &42.47          &\underline{35.95}          &38.26          &44.60          &48.04          &64.60          &62.99          &46.51 (1.06)            &47.94 (0.83) \\
DAML w/o Dst        &41.09          &44.22          &34.18          &37.47          &44.66          &48.74          &62.03          &62.52          &45.49 (2.43)            &48.23 (2.51) \\
DAML w/o Dir-Dst    &40.23          &42.92          &35.79  &\underline{38.64}          &45.05          &48.32          &64.97          &63.80          &46.51 (0.94)            &48.42 (1.07) \\
\hline
\end{tabular}}
\end{table*}

\subsubsection{Computational Cost}

\begin{figure*}[tb]
 \centering
 \includegraphics[scale=0.35]{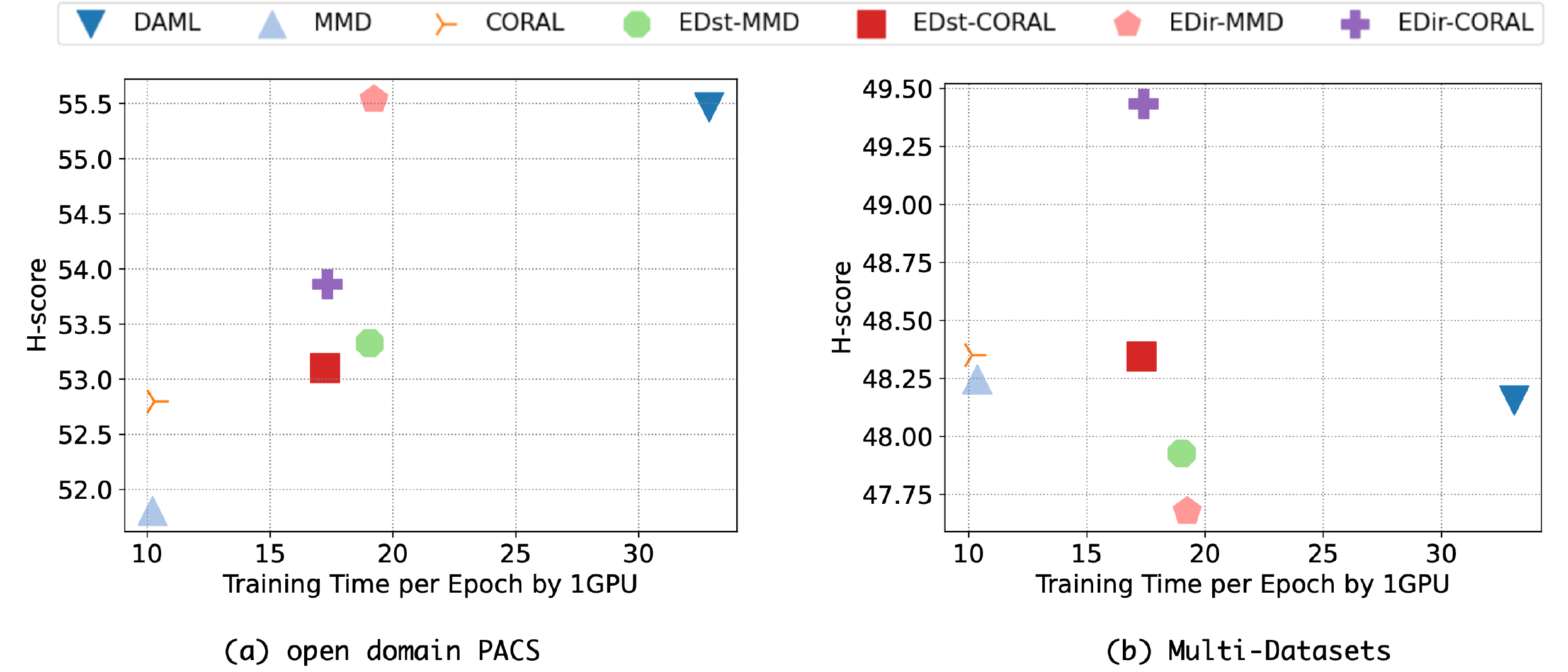}
 \caption{Scatterplot of H-score and training time for open-domain PACS and Multi-Datasets.}
\label{fig:h-score_vs_time}
\end{figure*}

Figure~\ref{fig:h-score_vs_time} compares the average training time per epoch of the variations of CORAL and MMD, and DAML. We used an NVIDIA A100 (80GB) GPU for model training. The tendency of training times is similar to that of open-domain Office-Home. For both open-domain PACS and Multi-Datasets, the training times of the variations of CORAL and MMD were less than that of DAML.

\end{document}